# Реализация экспертной системы для оценки инновационности технических решений

*В.К. Иванов [1], к.т.н., доцент, начальник управления информационных ресурсов и технологий, mtivk@mail.ru*
*И.В. Образцов [1], к.т.н., ведущий инженер по интеграции прикладных решений, sunspire@list.ru*
*Б.В. Палюх [1], д.т.н., профессор, зав. кафедрой информационных систем, pboris@tstu.tver.ru*

[1] *Тверской государственный технический университет, г. Тверь, 170026, Россия*

Представлено возможное решение задачи алгоритмизации количественной оценки показателей инновационности технических изделий, изобретений, технологий. Введены понятия технологической новизны, востребованности и имплементируемости – составных частей критерия инновационности продукта. Предложены модель и алгоритм вычисления каждого из указанных показателей инновационности в условиях неполноты и неточности, а иногда и противоречивости исходной информации.

В статье описывается разработанное специализированное ПО, которое является перспективным методологическим инструментом для использования интервальных оценок в соответствии с теорией свидетельств. Эти оценки применяются при анализе сложных многокомпонентных систем, агрегации больших объемов нечетких и неполных данных различной структуры. Представлены состав и структура мультиагентной экспертной системы, назначение которой – групповая обработка результатов измерений и оценок значений показателей инновационности объектов. Определяются активные элементы системы, их функциональность, роли, порядок взаимодействия, входные и выходные интерфейсы, общий алгоритм функционирования ПО. Описывается реализация программных модулей, приводится пример решения конкретной задачи по определению уровня инновационности технических изделий.

Разработанные подход, модели, методика и ПО могут быть использованы в реализации технологии хранилища характеристик объектов, обладающих значительным инновационным потенциалом. Формализация исходных данных задачи существенно повышает адаптивность предложенных методов к различным предметным областям. Появляется возможность обработки данных различной природы, полученных в результате опроса экспертов, из поисковой системы или даже с измерительного устройства, что способствует повышению практической значимости представленной разработки.

*Ключевые слова: инновационность, терм, хранилище данных, экспертная система, востребованность, имплементируемость, изобретение, оценка, свидетельство.*

Говоря о инновационности продуктов, необходимо учитывать три важных аспекта: что есть инновационность продукта, как определить количественную меру для ее оценки, с помощью каких удобных и эффективных инструментов можно вычислить уровень инновационности продукта и проанализировать результаты.

В этой связи актуальным представляется практическое решение задачи алгоритмизации количественной оценки показателей инновационности таких продуктов, как технические изделия, изобретения, технологии. В исследованиях, касающихся различных аспектов инновационного развития общества, определение «инновация» всегда основано на таких коннотациях, как «новый», «научный», «повышающий эффективность», «приносящий прибыль».

В данной работе введены понятия технологической новизны, востребованности и имплементируемости – составных частей критерия инновационности искомого объекта или продукта. Исходя из этих понятий разработаны модель и алгоритм вычисления каждого из указанных показателей инновационности в условиях неполноты и неточности, а иногда и противоречивости исходной информации. Реализацией предложенных концепций является описываемое в статье ПО специализированной экспертной системы, которая позволяет вычислить и проанализировать показатели инновационности продуктов, в том числе в динамике.

Предполагается, что разработанные подход, модели, методика и ПО могут быть использованы в реализации технологии хранилища характеристик следующих объектов:

– предприятия малого и среднего бизнеса (анализ ретроспективы и перспективы конкретных инноваций, поиск текущих и вероятных трендов в развитии инновационных товаров и услуг);

– бизнес-инкубаторы, стартапы (экспертиза инновационности проектных решений, оценка благоприятных для инвестиций факторов);

– образовательные учреждения (оценка инновационности образовательного контента, включая достижения обучающихся);

– органы управления (определение инновационных направлений для стимулирования деловой активности персонала, предприятий).

## Архитектура системы поддержки хранилища данных об инновациях

Типовая архитектура ПО включает слои представления, сервисов, бизнес-логики, доступа к данным, а также сквозную функциональность, которые должны обеспечивать взаимодействие пользователей и внешних систем с источниками данных (см., например, http://www.microsoft.com/architectureguide). На рисунке 1 приведена общая архитектура системы поддержки хранилища данных, заштрихованный блок – рассматриваемый в настоящей статье фрагмент системы. Коротко опишем состав и назначение основных компонентов:

– специализированные прикладные системы для информационного обеспечения имплементации инноваций;

– внешние (по отношению к рассматриваемой) *системы поддержки принятия решений* (СППР);

– визуализация состава хранилища данных об инновациях, поисковых паттернов и результатов поиска по ним, индикаторов инновационности объектов хранилища, наборов связанных объектов;

– программные интерфейсы для взаимодействия с внешними СППР и компонентами слоя представления;

– активные элементы, или акторы (агенты); детализация каждого типа акторов представлена в таблице 1;

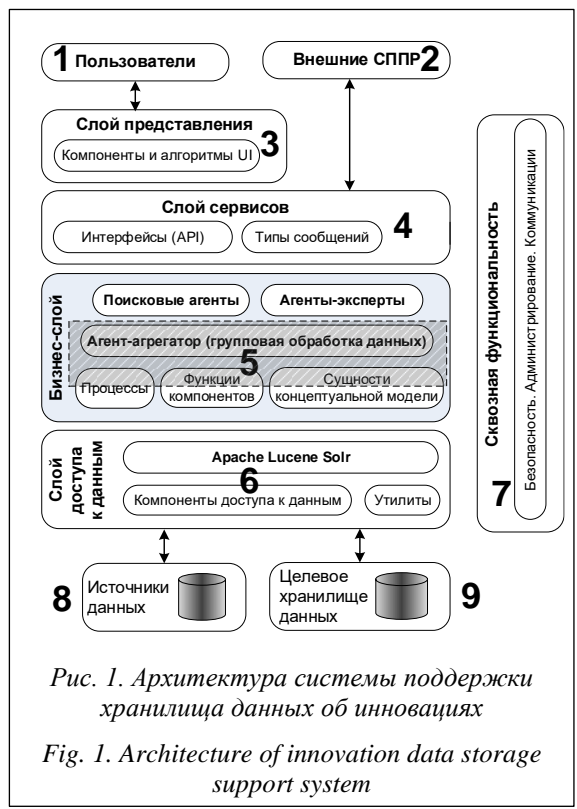

*Рис. 1. Архитектура системы поддержки хранилища данных об инновациях*

*Fig. 1. Architecture of innovation data storage support system*

– компонент Apache Lucene Solr – программная реализация модели векторного пространства документов для предварительной обработки данных, планируемых к помещению в целевое хранилище данных;

– компоненты, отвечающие за безопасность данных, администрирование, сетевые коммуникации;

– ресурсы Интернета, специализированные хранилища и БД;

– реестр инновационных решений научно-технических задач.

## Показатели инновационности объектов

Введем понятия технологической новизны, востребованности и имплементируемости как составных частей критерия инновационности искомого объекта. Количественная оценка этих показателей основана на гипотезе об адекватности отображения жизненного цикла продуктов в различных хранилищах данных при условии доступа к достаточному количеству таких хранилищ. Для поиска информации о потенциально инновационных объектах в хранилищах

*Таблица 1*

**Содержания процессов, функций компонентов и сущностей концептуальной модели
бизнес-слоя**

*Table 1*

**Contents of processes, functions of components and entities of the business layer conceptual model**

| Агент | Бизнес-процессы | Бизнес-компоненты | Бизнес-сущности |
|---|---|---|---|
| Агент-гене-ратор ис-ходных данных (АГИД) | Эволюционное формирова-ние лингвистической мо-дели архетипа объекта и эф-фективного мультимноже-ства поисковых запросов | Генетический алгоритм для получения эффективного мультимножества поиско-вых запросов. Алгоритм расчета фитнес-функции и фильтрации результатов по-иска | Лингвистическая модель архетипа искомого объ-екта. Поисковый паттерн. Общий понятийный базис |
| Поисковые агенты (ПА) | Поиск информации об инно-вационных решениях в гете-рогенных хранилищах и БД. Организация взаимодей-ствия интеллектуальных по-исковых агентов | Генетический алгоритм для получения эффективного мультимножества поиско-вых запросов. Алгоритм расчета фитнес-функции и фильтрации результатов поиска | Лингвистическая модель архетипа искомого объ-екта. Поисковый паттерн. Параметры для расчета ин-дикаторов инновационно-сти объектов. Общий поня-тийный базис |
| Агенты-эксперты (АЭ) | Экспертиза целевых объек-тов в прикладных областях | Собственные методики экспертов | Индикаторы инновацион-ности объектов хранилища. Нечеткие показатели веро-ятности заданных свойств инновационности. Общий понятийный базис |
| Агент-агре-гатор (АА) | Обработка результатов из-мерений исходных данных для расчета вероятностных значений показателей инно-вационности объектов, в том числе полученных из нескольких источников с учетом надежности источ-ника. Вычисление функций доверия и правдоподобия (теория свидетельств) | Модель вычислений инди-каторов инновационности объектов. Алгоритм и мето-дика групповой обработки результатов измерений уровня инновационности объекта | Индикаторы инновацион-ности объектов хранилища. Нечеткие показатели веро-ятности заданных свойств инновационности: функ-ции доверия и правдоподо-бия. Общий понятийный базис |

данных предложена лингвистическая модель архетипа искомого объекта, образующая поис-ковый паттерн. Термы модели классифициру-ются как ключевые свойства, описывающие структуру объекта, условия применения или результаты функционирования. Область опре-деления архетипа определяется маркером. По-исковые запросы конструируются как комби-нации термов и маркера. При значительном ко-личестве термов используется генетический алгоритм генерации запросов и фильтрации ре-зультатов, позволяющий получить квазиопти-мальный набор поисковых запросов [1]. Алго-ритм использует принципы работы с метаэври-стиками, применяемыми для решения задач стохастической оптимизации [2].

Под технологической новизной архетипа объекта понимаются его значительные улуч-шения, новый способ использования или предоставления (субъектами новизны явля-ются потенциальные пользователи или сам производитель). Оценка индикатора новизны основана на нормированном интегральном зна-чении числа найденных элементов информа-ции об объекте (документов, записей и т.п.) в результате поиска в гетерогенных БД. Пред-полагается, что для новых объектов количество найденных элементов информации, релевант-ных поисковому паттерну, будет меньше, чем для давно существующих и известных объек-тов. Формальное выражение для вычисления значения индикатора новизны объекта следую-щее:

$$Nov = 1 - \frac{1}{S}\sum_{k=1}^{S} R_{01}(R_k, \ldots) , \qquad (1)$$

где $Nov$ – новизна архетипа объекта; $S$ – общее количество выполненных запросов к БД; $R_k$ – число документов, найденных в результате выполнения $k$-го запроса к БД, содержащей информацию о рассматриваемой предметной области; $R_{01}(R_k, \ldots)$ – вариативная функция, нормирующая значение $R_k$ на диапазон [0; 1]. Могут быть использованы различные функции $R_{01}(R_k, \ldots)$. Например:

– линейное нормирование

$$R_{01} = \frac{R_k - R_{min}}{R_{max} - R_{min}} ; \qquad (2)$$

– нелинейное нормирование со статистическими характеристиками данных

$$R_{01} = \frac{R_k - \overline{R}}{\sigma} ; \qquad (3)$$

– нелинейное экспоненциальное нормирование

$$R_{01} = 1 - \exp(1 - \frac{R_k}{R_{min}}) . \qquad (4)$$

Здесь $R_{min}$ и $R_{max}$ – наименьшее и наибольшее, соответственно, число документов, найденных при выполнении всех $S$ запросов; $\overline{R}$ – среднее число документов, найденных при выполнении всех $S$ запросов; $\sigma$ – дисперсия $R_k$.

Линейное нормирование (2) предпочтительно, когда значения $R_k$ достаточно равномерно заполняют интервал $R_{min}$–$R_{max}$. Если среди всех значений $R_k$ имеются редкие аномалии, намного превышающие типичный разброс, следует использовать выражение (3), а если $R_{max} \to \infty$, целесообразно применить выражение (4).

Востребованность архетипа объекта – это осознанная потенциальным производителем необходимость в этом объекте, оформленная в спрос. Оценка индикатора востребованности основана на нормированном интегральном значении частоты обращения пользователей к информации о потенциально инновационном продукте или услуге.

Формальное выражение для вычисления значения индикатора новизны объекта:

$$Rel = \frac{1}{S}\sum_{k=1}^{S} F_{01}(F_k, \ldots) , \qquad (5)$$

где $Rel$ – востребованность архетипа объекта; $F_k$ – частота обращения пользователей к информации о потенциально инновационном продукте или услуге, найденной при выполне-нии $k$-го запроса из $S$ запросов; $F_{01}(F_k, \ldots)$ – вариативная функция, нормирующая значение $F_k$ на диапазон [0; 1]. Выбор вида функции $F_{01}(F_k, \ldots)$ осуществляется аналогично выбору функции $R_{01}(R_k, \ldots)$ при расчете $Nov$ в соответствии с (2)–(4). Используемые в качестве аргументов $F_{min}$, $F_{max}$, $\overline{F}$ и $\sigma$ обозначают наименьшее, наибольшее и среднее значения, а также дисперсию $F_k$.

Имплементируемость архетипа объекта определяет технологическую обоснованность, физическую осуществимость и способность интеграции объекта в систему для получения желаемого эффекта. Оценка индикатора имплементируемости основана на нормированном значении среднего периода восстановления уровня новизны и/или востребованности архетипа объекта. Предполагается, что со временем архетип объекта теряет новизну и/или востребованность. Однако новые технологии, конструкция, улучшенные функциональные и потребительские характеристики могут повысить значения индикаторов новизны и востребованности. Чем быстрее это происходит, тем выше имплементируемость.

Формальное выражение для вычисления значения индикатора имплементируемости объекта:

$$Imp = 1 - \frac{1}{2}\left(\overline{LN}_{01}^{max}(Nov(t)) + \overline{LR}_{01}^{max}(Rel(t))\right), (6)$$

где $Imp$ – имплементируемость архетипа объекта; $Nov(t)$ – функция, показывающая зависимость новизны архетипа объекта от времени и определенная на временном интервале $[t_0; t_m]$; $Rel(t)$ – функция зависимости востребованности архетипа объекта от времени, определенная на том же временном интервале $[t_0; t_m]$; $\overline{LN}_{01}^{max}$ и $\overline{LR}_{01}^{max}$ – средние расстояния между двумя последовательными точками временного ряда $t_i$, $t_{i+1} \in [t_0; t_m]$ локальных максимумов функций $Nov(t)$ и $Rel(t)$ соответственно.

Учитывая непосредственную количественную оценку индикаторов инновационности, будем считать, что этот подход дополняет традиционные [3, 4].

## Обоснование использования теории свидетельств

Поскольку возможны очевидные неполнота и неточность информации об объектах, полученной из различных источников, введены нечеткие показатели вероятности того, что объ-ект обладает технологической новизной, востребован потребителями и реализуем, то есть может быть имплементирован.

В экспертных системах в случаях неопределенности исходных данных выбор того или иного математического метода обработки зависит от степени такой неопределенности. Вероятностные и статистические методы, методы нечеткой логики применяются в условиях частичной неопределенности и требуют обработки больших объемов информации, оперирования с повторными выборками, детерминированного характера вероятностных характеристик [5–7]. Для смягчения подобных требований целесообразно применение теории свидетельств Демпстера–Шафера [8, 9]. Математический аппарат теории свидетельств является распространенным современным подходом, применяемым в различных задачах оптимизации, диагностики технических систем, оценки уровней соответствия технических показателей целевым значениям, оценки уровня инвестиционного потенциала технических решений и инноваций и в других актуальных задачах.

Принимается, что базовая вероятность $m$ попадания результатов измерения показателя инновационности объекта ($Nov, Rel$ и $Imp$) в интервал значений $A$ определяется следующим образом:

$$m : P(\Omega) \to [0;1], m(\varnothing) = 0, \sum_{A \in P(\Omega)} m(A) = 1 , \quad (7)$$

где $\Omega$ – множество значений результатов измерения показателя; $P(\Omega)$ – множество всех подмножеств $\Omega$.

Далее для заданных $k$ интервалов рассчитываются функция доверия: $Bel(A) = \sum_{A_k : A_k \subseteq A} m(A_k)$

и функция правдоподобия: $Pl(A) = \sum_{A_k : A_k \cap A} m(A_k)$,

которые определяют верхнюю и нижнюю границы вероятности обладания объектом заданного свойства. Таким образом, дается оценка значениям показателей $Nov, Rel$ и $Imp$ в условиях неполноты и неточности информации об объектах.

Математический аппарат теории свидетельств ориентирован на получение объективной модели согласования экспертных суждений, наблюдений или измерений. Условия отсутствия знаний об объекте, о предыстории процесса, невозможность использования повторных выборок позволяют применять теорию свидетельств к задачам, в которых теория вероятностей и нечеткая логика принципиально неприменимы. В целях повышения гибкости и достоверности получаемых моделей в ряде исследований предлагаются различные модификации математического аппарата теории свидетельств в части правил агрегирования свидетельств из независимых источников и учета конфликтности экспертных мнений [10–13].

На предыдущих этапах работы по данному проекту изложены принципы специфического применения теории свидетельств к оценке показателей инновационности объектов, описаны лингвистическая модель, технология обработки результатов измерений, полученных из поисковых систем, а также разработана методика оценки достоверности источника данных. В [14] как объект исследования рассмотрена электронная информационно-образовательная среда университета в контексте кооперации образовательных и бизнес-процессов. Описаны возможности применения предложенных ранее подходов к оценке новизны, востребованности и имплементируемости компонентов среды. В работе [15] отражена концепция практического применения теории свидетельств для оценки инновационного потенциала технических решений и изобретений, а также моделирования, диагностики и оценки состояния сложных производственных систем.

## Алгоритмизация ПО оценки инноваций

Общее представление функциональности ПО системы показано на рисунке 2 в виде диаграммы сценариев UML.

Поведение системы в течение конкретного сеанса работы представлено на диаграмме последовательностей UML (рис. 3). Изображена последовательность сообщений между взаимодействующими акторами и программными компонентами. На рисунке 3 использованы следующие обозначения: 1 – лингвистическая модель архетипа объекта (поисковый паттерн) и мультимножество поисковых запросов; 2 – параметры для расчёта индикаторов инновационности объектов и поисковый паттерн; 3 – уточнение поискового паттерна; 4 – индикаторы инновационности объектов (экспертное мнение); 5 – уточнение запроса эксперту; 6 – рассчитанные индикаторы инновационности объектов (аналоги экспертных мнений); 7 – комбинация свидетельств (индикаторов инновационности ) с учетом надёжности источника;

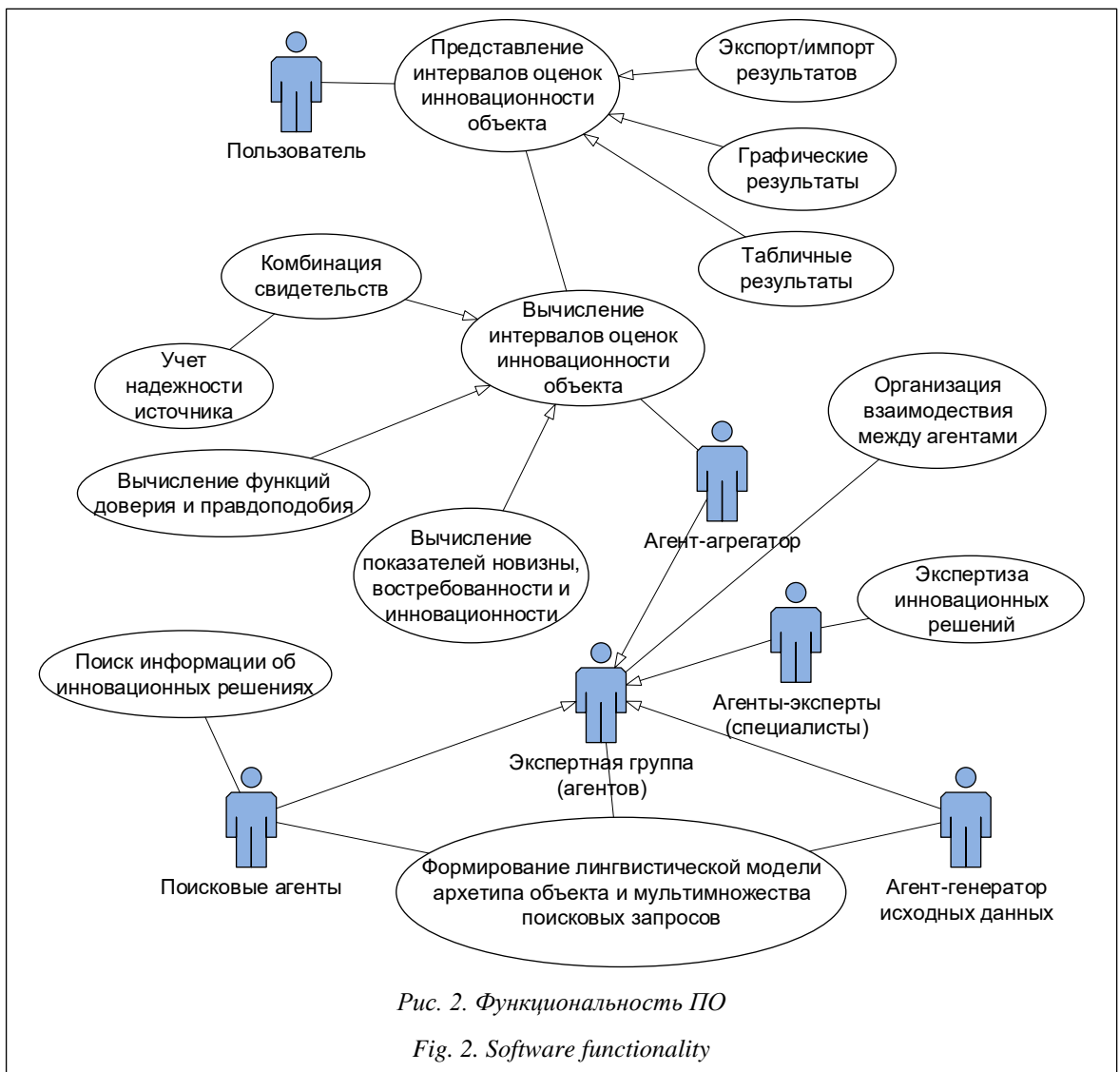

*Рис. 2. Функциональность ПО*

*Fig. 2. Software functionality*

8 – значения функций доверия и правдоподобия; 9 – интервалы оценок инновационности объекта; 10 – табличные результаты оценки инновационности объекта; 11 – графические результаты оценки инновационности объекта; 12 – данные для экспорта результатов оценки инновационности объекта.

Опишем алгоритм вычисления количественной оценки инновационности многокомпонентного объекта (рис. 4). Структуризация исходных данных происходит по четырем абстрактным категориям – компоненты, показатели, экспертные группы, оценки. После сбора исходных данных (действие 1) осуществляется сортировка экспертных мнений и их аналогов (оценок) с последующим формированием таблицы свидетельств (действие 2). Если в пределах одной экспертной группы встречаются одинаковые оценки, они объединяются в одно

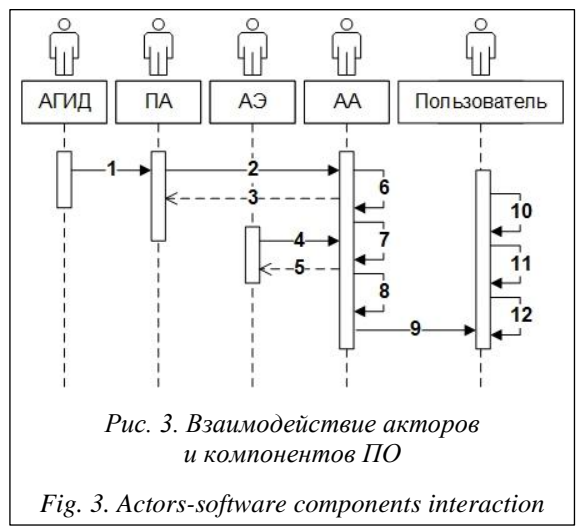

*Рис. 3. Взаимодействие акторов*

*и компонентов ПО*

*Fig. 3. Actors-software components interaction*

свидетельство с учетом количества экспертов, давших оценку. Далее производится расчет базовых вероятностей для каждого свидетельства. Обработка свидетельств выполняется по всем рассматриваемым показателям для каждого компонента объекта. Цикл формирования таблицы свидетельств (действия 2 и 3) заканчивается при достижении последней экспертной оценки (действие 3). Далее выполняются сортировка (действие 4) и комбинирование (действие 6) свидетельств из различных источников. Целью является объединение свидетельств, числовые интервалы которых пересекаются друг с другом, с учетом влияния конфликтных свидетельств, числовые интервалы которых не пересекаются. Процедура комбинирования объединена с расчетом границ математического ожидания интервалов и значений функций доверия и правдоподобия. Цикл комбинирования свидетельств (действия 4–6) заканчивается после обработки всех свидетельств (действие 5). Алгоритм завершается ранжированием (действие 7) комбинированных оценок. В таблицу интегральных оценок выводятся оценки, для которых нижняя и верхняя границы вероятности наличия свойства инно-

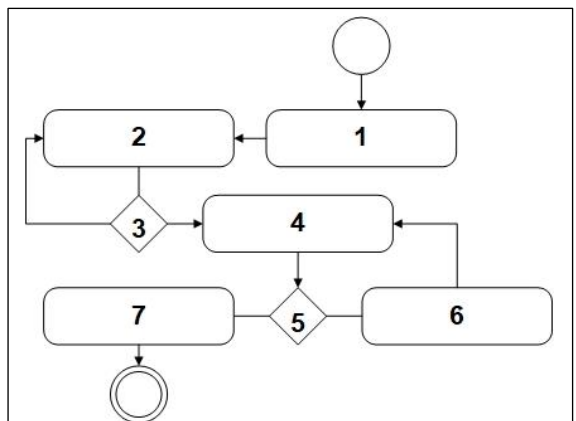

*Рис. 4. Алгоритм вычисления оценки
инновационности объекта*

*Fig. 4. Algorithm for computing the assessment
of the innovativeness of an object*

вационности у объекта максимальны.

Отметим, что в контексте рассматриваемой задачи понятия «экспертная группа», «эксперт», «экспертное мнение» являются абстрактными и могут интерпретироваться в отношении не только экспертов-специалистов, но и поисковых агентов, устройств измерений, результатов наблюдений и/или вычислений.

## Реализация ПО

Прототип экспертной системы реализован в виде веб-приложения с графическим интерфейсом пользователя. Приложение разработано на языке SpiderBasic, компилятор которого генерирует оптимизированный код JavaScript для выполнения в браузере с поддержкой стандарта HTML5.

Графический интерфейс приложения имеет иерархическую структуру и состоит из главного окна (см. http://www.swsys.ru/uploaded/image/2019-4/2019-4-dop/14.jpg) и ряда вспомогательных диалоговых окон, отображаемых в процессе работы. Взаимодействие пользователя с приложением сегментировано на раздельные операции посредством переключаемых объектных контейнеров, каждый из которых содержит группы стандартных элементов управления – текстовые поля, электронные таблицы, списки, кнопки, переключатели и флажки. С помощью графического интерфейса пользователь последовательно настраивает параметры экспертной системы, заполняет опросный лист, а также получает доступ к таблицам свидетельств и выходным расчетным показателям.

Представим основные классы данных, используемые в экспертной системе.

*Класс Estimate – Оценка*: Lingvo.s – лингвистический терм (string), LBound.f – нижняя граница интервала (float), UBound.f – верхняя граница интервала (float).

*Класс ExpGrp – экспертная группа*: GroupName.s – название экспертной группы, ExperCount.i – количество экспертов в группе.

*Класс Evidence – свидетельство*: Lingvo.s – лингвистический терм (string), LBound.f – нижняя граница интервала (float), UBound.f – верхняя граница интервала (float), Numb_c.i – количество экспертов, давших оценку (integer), Numb_N.i – количество экспертов в группе (integer), Val_mA.f – базовая вероятность или масса (float), Val_bA.f – функция доверия (float), Val_pA.f – функция правдоподобия (float).

*Класс DSstruc – структура Демпстера–Шейфера*: LBound.f – нижняя граница (float), UBound.f – верхняя граница (float), Massa.f – масса (float).

*Класс MatrAB – матрица пересечения множеств*: SubM.f – произведение масс (float), Stat.i – статус пересечения интервалов (integer), SumL.f – объединенная нижняя граница (float), SumU.f – объединенная верхняя граница (float).

*Класс SourceData – исходные данные*: ComponentNumber.i – количество компонентов, IndicatorNumber.i – количество показателей, ExpGroupsNumber.i – количество экспертных групп, EstimatesNumber.i – количество оценок в шкале, RoundDigsNumber.i – количество разрядов округления, InterviewNumber.i – количество результатов опроса, List ComponentNames.s() – список названий компонентов, List IndicatorNames.s() – список названий показателей, List ExpertGroupes.ExpGrp() – список экспертных групп с двумя подтипами (название группы, количество человек), List EstimateScale.estimate() – список оценок шкалы с тремя подтипами (лингвистический терм, нижняя граница, верхняя граница), List InterviewRslt. estimate() – список результатов опроса экспертов с тремя подтипами (лингвистический терм, нижняя граница, верхняя граница).

Данные конкретного расчета показателей инновационности могут быть экспортированы в файл в формате JSON. При необходимости дальнейшей обработки эти данные могут быть импортированы в экспертную систему. Стандартизация формата данных приложения особенно актуальна при оценке показателей на основе большого количества технических измерений или расчетов.

Приведем фрагмент файла экспорта данных, построенного по результатам выполнения поисковых запросов в нескольких поисковых системах с целью оценки изобретений и технических решений по нормированному показателю технологической новизны:

```
{
    "ComponentNumber": 10,
    "IndicatorNumber": 1,
    "ExpGroupsNumber": 5,
    "EstimatesNumber": 20,
    "RoundDigsNumber": 3,
    "InterviewNumber": 800,
    "ComponentNames": ["Электрический глаз",
"Ген-активированный материал для регенерации тканей", "Имплантация миниконтура", "Лечение пародонта", "Электронный индикатор уровня"],
    "IndicatorNames": ["Нормированный показатель КН4"],
    "ExpertGroupes": [{"GroupName": "Yandex", "ExperCount": 16}, {"GroupName": "ЕГИСУ НИОКТР", "ExperCount": 16}, {"GroupName": "Google", "ExperCount": 16}],
    "EstimateScale":      [{"Lingvo":    "0–5 %",
"LBound": 0, "UBound": 0.05}, {"Lingvo": "95–100%", "LBound": 0.95, "UBound": 1}],
```

```
    "InterviewRslt":   [{"Lingvo":   "95–100 %",
"LBound": 0.95, "UBound": 1}, {"Lingvo": "60–65 %", "LBound": 0.6, "UBound":           0.65},
{"Lingvo": "0–5 %", "LBound": 0, "UBound": 0.05}]
}
```

### Пример работы экспертной системы

Рассмотрим обобщенно оценку показателя инновационности с помощью обсуждаемой экспертной системы. Зададим оценочную интервальную числовую шкалу от 0 до 1,0, каждому интервалу присвоим соответствующий лингвистический терм. Как правило, эксперту-специалисту удобнее работать с лингвистическими термами, а числовая составляющая шкалы необходима для математической обработки данных.

В случае использования результатов измерений в качестве экспертных данных числовые значения (получаемые, например, от поисковых агентов) следует соотносить с соответствующими им числовыми интервалами оценочной шкалы. Причем назначение узких оценочных интервалов дает более точные выходные значения, но приводит к повышению вероятности конфликтов экспертных оценок. В свою очередь, использование широких оценочных интервалов снижает вероятность конфликтности экспертных оценок, но при этом приводит и к снижению точности выходных данных экспертной системы.

Рассмотрим ситуацию, в которой присутствуют три независимых источника свидетельств – *A*, *B* и *C*. Эти источники в терминах предлагаемой экспертной оценки представляют собой экспертные группы со строго определенным количеством экспертов. Предположим, что экспертные группы *A*, *B* и *C* включают 120, 80 и 50 экспертов соответственно. Примеры оценок экспертных групп приведены в таблице 2.

Каждая экспертная оценка представляет собой уникальное свидетельство в пределах своей группы. Основной характеристикой свидетельств является его базовая вероятность или масса $m(A)$, соотнесенная с границами числового интервала в исходной оценочной шкале:

$$M(A_i) = A_i)/N_i, \qquad (8)$$

где $A_i$ – интервал *i*-го свидетельства; $A_i)$ – количество экспертов, давших *i*-е свидетельство; $N_i$ – общее количество экспертов в группе.

Необходимо скомбинировать свидетельства из экспертных групп *A*, *B* и *C* между собой,

### Примеры оценок экспертных групп

### Examples of expert groups' evaluations

| № п/п | Количество экспертов | Лингвистический терм оценки | Нижняя граница числового интервала | Верхняя граница числового интервала |
|---|---|---|---|---|
| | | Результаты опроса группы A | | |
| 1 | 10 | основная № 1 | 0,00 | 0,33 |
| 2 | 5 | основная № 2 | 0,34 | 0,66 |
| 3 | 10 | вспомогательная № 9 | 0,89 | 1,00 |
| 4 | 20 | основная № 3 | 0,67 | 1,00 |
| 5 | 5 | вспомогательная № 8 | 0,78 | 0,88 |
| 6 | 15 | вспомогательная № 3 | 0,23 | 0,33 |
| 7 | 15 | вспомогательная № 5 | 0,45 | 0,55 |
| 8 | 5 | вспомогательная № 2 | 0,12 | 0,22 |
| 9 | 15 | вспомогательная № 7 | 0,67 | 0,77 |
| 10 | 5 | вспомогательная № 4 | 0,34 | 0,44 |
| 11 | 5 | вспомогательная № 6 | 0,56 | 0,66 |
| 12 | 5 | вспомогательная № 1 | 0,00 | 0,11 |
| 13 | 5 | нулевая оценка | 0,00 | 0,00 |
| Всего: | 120 | | | |
| | | Результаты опроса экспертов группы B | | |
| 1 | 10 | вспомогательная № 4 | 0,34 | 0,44 |
| 2 | 5 | вспомогательная № 8 | 0,78 | 0,88 |
| 3 | 20 | основная № 2 | 0,34 | 0,66 |
| 4 | 15 | основная № 1 | 0,00 | 0,33 |
| 5 | 5 | вспомогательная № 5 | 0,45 | 0,55 |
| 6 | 20 | основная № 3 | 0,67 | 1,00 |
| 7 | 5 | вспомогательная № 9 | 0,89 | 1,00 |
| Всего: | 80 | | | |
| | | Результаты опроса экспертов группы C | | |
| 1 | 10 | вспомогательная № 9 | 0,89 | 1,00 |
| 2 | 5 | вспомогательная № 1 | 0,00 | 0,11 |
| 3 | 5 | вспомогательная № 7 | 0,67 | 0,77 |
| 4 | 5 | вспомогательная № 6 | 0,56 | 0,66 |
| 5 | 10 | нулевая оценка | 0,00 | 0,00 |
| 6 | 5 | основная № 2 | 0,34 | 0,66 |
| 7 | 5 | основная № 3 | 0,67 | 1,00 |
| 8 | 5 | вспомогательная № 8 | 0,78 | 0,88 |
| Всего: | 50 | | | |

учитывая их однозначность и конфликтность. В разработанном алгоритме используется правило комбинирования Демпстера, суть которого заключается в формировании матрицы пересечения числовых интервалов свидетельств и расчете комбинированных базовых вероятностей, нормализованных с помощью коэффициента конфликтности. В таблице 3 для примера представлена матрица пересечения числовых интервалов свидетельств из группы *A* (все свидетельства) и группы *B* (свидетельства из диапазонов основных оценок № 1 и № 2). Номера интервалов групп *A* и *B* обозначены как *i*(*A*) и

*j*(*B*) соответственно. В ячейках матрицы представлены произведения базовых вероятностей свидетельств. Жирным шрифтом выделены значения в ячейках, соответствующие областям полного или частичного пересечения числовых интервалов свидетельств. Остальные ячейки соответствуют областям конфликтующих (непересекающихся) свидетельств.

Базовые вероятности пар неконфликтных свидетельств, имеющих одинаковые числовые интервалы, суммируются:

$$\sum m_{AB} = \sum_{A_i^{(A)} \cap A_j^{(B)} = A} m_A(A_i^{(A)}) \cdot m_B(A_j^{(B)}). \qquad (9)$$

Базовые вероятности пар конфликтных свидетельств, числовые интервалы которых не пересекаются, также суммируются:

$$\sum m_{AB}^{\varnothing} = \sum_{A_i^{(A)} \cap A_j^{(B)} = \varnothing} m_A(A_i^{(A)}) \cdot m_B(A_j^{(B)}).  \quad (10)$$

Для данного примера $\sum m_{AB} = 0{,}2787063$ и $\sum m_{AB}^{\varnothing} = 0{,}7214938$. Учет конфликтности свидетельств осуществляется с помощью константы нормализации: $K = 1/1 - \sum m_{AB}^{\varnothing} = 3{,}5905837$. Рассчитываются нормализованные базовые вероятности неконфликтных свидетельств для объединенных интервалов:

«вспомогательная № 9». Для определения наиболее вероятных оценок необходимо рассчитать значения функций доверия $Bel(A)$ и правдоподобия $Pl(A)$; результаты расчета представлены в таблице 4. Из них следует, что наиболее вероятна оценка с интервалом 0,67–1,00, соответствующая лингвистическому терму «основная № 3» в исходной оценочной шкале.

В процессе работы экспертной системы строится диаграмма обобщенных аккумулированных функций распределения, визуализирующая числовые интервалы исходных свидетельств до процедуры комбинирования и их ба-

*Таблица 3*

**Матрица пересечения числовых интервалов для групп A и B (пример)**

*Table 3*

**Numerical intervals intersection matrix (example) for groups A and B**

| | $i(A)$ | 1 | 2 | 3 | 4 | 5 | 6 | 7 | 8 | 9 | 10 | 11 | 12 | 13 |
|---|---|---|---|---|---|---|---|---|---|---|---|---|---|---|
| $j(B)$ | $M(A_i)$ | 0,0833 | 0,0417 | 0,0833 | 0,1667 | 0,0417 | 0,125 | 0,125 | 0,0417 | 0,125 | 0,0417 | 0,0417 | 0,0417 | 0,0417 |
| 3 | 0,2500 | 0,020825 | **0,010425** | 0,020825 | 0,041675 | 0,010425 | 0,03125 | **0,03125** | 0,010425 | 0,03125 | **0,010425** | **0,010425** | 0,010425 | 0,010425 |
| 4 | 0,1875 | **0,015619** | 0,007819 | 0,015619 | 0,031256 | 0,007819 | **0,023438** | 0,023438 | **0,007819** | 0,023438 | 0,007819 | 0,007819 | **0,007819** | **0,007819** |

*Таблица 4*

**Значения функций доверия и правдоподобия**

*Table 4*

**Values of Trust and Likelihood Functions**

| Нижняя граница $L$ объединенного интервала | Верхняя граница $U$ объединенного интервала | Значение функции доверия $Bel(A)$ | Значение функции правдоподобия $Pl(A)$ |
|---|---|---|---|
| 0,00 | 0,33 | 0,1900 | 0,1900 |
| 0,34 | 0,66 | 0,1557 | 0,1557 |
| 0,67 | 1,00 | 0,6544 | 0,6544 |
| 0,89 | 1,00 | 0,0106 | 0,6517 |
| 0,78 | 0,88 | 0,0026 | 0,6438 |

$$\sum m_{AB}(A) = K \cdot \sum m_{AB}.  \quad (11)$$

Далее необходимо произвести комбинирование полученных объединенных свидетельств условного источника $AB$ со свидетельствами экспертной группы $C$. Порядок вычислений аналогичен приведенному для групп $A$ и $B$. После выполнения процедуры комбинирования получены пять производных свидетельств, числовые интервалы которых соответствуют оценкам: «основная № 1», «основная № 2», «основная № 3», «вспомогательная № 8» и зовые вероятности (см. http://www.swsys.ru/uploaded/image/2019-4/2019-4-dop/15.jpg).

По оси абсцисс диаграммы откладываются числовые интервалы фокальных элементов в порядке возрастания, а по оси ординат – значения масс фокальных элементов с накоплением. Данные диаграммы наглядно иллюстрируют согласованность (наложение областей диаграммы) и конфликтность экспертных мнений из нескольких экспертных групп. Разными цветами обозначены свидетельства из исходных экспертных групп. Области, в которых все три цвета накладываются друг на друга, соответствуют областям согласованности экспертных мнений. Ширина столбцов диаграммы соответствует числовым интервалам оценочной шкалы. Кроме того, диаграмма отражает оценки, которые не были задействованы в экспертном опросе (горизонтальные линии между заполненными областями).

Демонстрационная версия приложения доступна по адресу http://virtlabs.tech/apps/DST/Dempster_Shafer_App.html. Функциональность приложения поддерживается для всех популярных веб-браузеров.

## Заключение

Представляется, что описанное в статье ПО является перспективным методологическим инструментом и в определенной степени спо-собствует решению таких актуальных задач, как имплементация алгоритма количественной оценки инновационных свойств объектов, использование интервальных оценок в соответствии с теорией свидетельств при анализе сложных многокомпонентных систем, агрегация больших объемов нечетких и неполных данных различной структуры.

Важной особенностью представленной разработки является возможность абстрактной формализации исходных данных задачи, что существенно повышает адаптивность данного метода к различным предметным областям. Таким образом, появляется возможность обработки данных различной природы, полученных в результате опроса экспертов, из поисковой системы или даже с измерительного устройства, что, в свою очередь, способствует повышению практической значимости представленной разработки.

## Implementing an expert system to evaluate technical solutions innovativeness

***V.K. Ivanov*** [1], *Ph.D. (Engineering), Associate Professor, Head of Information Resources and Technologies Office, mtivk@mail.ru*
***I.V. Obraztsov*** [1], *Ph.D. (Engineering), Leading Application Integration Engineer, sunspire@list.ru*
***B.V. Palyukh*** [1], *Dr.Sc. (Engineering), Professor, Head of Information System Department, pboris@tstu.tver.ru*

[1] *Tver State Technical University, Tver, 170026, Russian Federation*

**Abstract.** The paper presents a possible solution to the problem of algorithmization for quantifying innovativeness indicators of technical products, inventions and technologies. The concepts of technological novelty, relevance and implementability as components of product innovation criterion are introduced. Authors propose a model and algorithm to calculate every of these indicators of innovativeness under conditions of incompleteness and inaccuracy, and sometimes inconsistency of the initial information.

The paper describes the developed specialized software that is a promising methodological tool for using interval estimations in accordance with the theory of evidence. These estimations are used in the analysis of complex multicomponent systems, aggregations of large volumes of fuzzy and incomplete data of various structures. Composition and structure of a multi-agent expert system are presented. The purpose of such system is to process groups of measurement results and to estimate indicators values of objects innovativeness. The paper defines active elements of the system, their functionality, roles, interaction order, input and output interfaces, as well as the general software functioning algorithm. It describes implementation of software modules and gives an example of solving a specific problem to determine the level of technical products innovation.

The developed approach, models, methodology and software can be used to implement the storage technology to store the characteristics of objects with significant innovative potential. Formalization of the task's initial data significantly increases the possibility to adapt the proposed methods to various subject areas. There appears an opportunity to process data of various natures, obtained during experts' surveys, from a search system or even a measuring device, which helps to increase the practical significance of the presented research.

**Keywords:** innovation, term, data storage, expert system, relevance, implementability, invention, evaluation, certificate.

*Acknowledgements. The research was financially supported by RFBR, projects no. 18-07-00358, 17-07-01339.*